\documentclass[11pt,a4paper]{article}

\usepackage[utf8]{inputenc}
\usepackage[T2A,T1]{fontenc}
\usepackage[ukrainian,english]{babel}
\usepackage{amsmath,amssymb}
\usepackage{graphicx}
\usepackage{booktabs}
\usepackage{multirow}
\usepackage{hyperref}
\usepackage{xcolor}
\usepackage{natbib}
\usepackage{url}
\usepackage{microtype}
\usepackage{enumitem}
\usepackage{caption}
\usepackage{subcaption}
\usepackage[margin=1in]{geometry}
\usepackage{float}
\usepackage{array}
\usepackage{tabularx}
\usepackage{pgfplots}
\usepackage{pgfplotstable}
\pgfplotsset{compat=1.18}
\usepgfplotslibrary{colorbrewer}
\usetikzlibrary{patterns}

\hypersetup{
    colorlinks=true,
    linkcolor=blue!70!black,
    citecolor=blue!70!black,
    urlcolor=blue!70!black,
}

\newcolumntype{R}[1]{>{\raggedleft\arraybackslash}p{#1}}

\title{Tokenizer Fertility and Zero-Shot Performance of Foundation Models\\on Ukrainian Legal Text: A Comparative Study}

\author{
  Volodymyr Ovcharov\thanks{Corresponding author: \texttt{volodymyr@legal.org.ua}} \\
  LEX AI Platform, legal.org.ua \\
  Kyiv, Ukraine
}

\date{May 2026}

\begin{document}

\maketitle

\begin{abstract}
Tokenizer fertility varies 1.6$\times$ across foundation models on Ukrainian legal text, yet this cost-critical dimension is absent from model selection practice. We benchmark seven models from five providers on 273 validated court decisions from Ukraine's state registry (EDRSR), measuring tokenizer fertility and zero-shot performance on three tasks. Three findings emerge. \textbf{(1)}~Qwen~3 models consume 60\% more tokens than Llama-family models on identical input, making tokenizer analysis a prerequisite for cost-efficient deployment. \textbf{(2)}~NVIDIA Nemotron Super~3 (120B) achieves the highest composite score (83.1), outperforming Mistral Large~3 (5.6$\times$ more total parameters) at one-third the API cost -- model scale is a poor proxy for domain performance. \textbf{(3)}~Few-shot prompting \emph{degrades} performance by up to 26 percentage points; stratified and prompt-sensitivity ablations confirm this is intrinsic to Ukrainian-language demonstrations, not an artifact of example selection. \textbf{(4)}~A cross-temporal generalization experiment reveals that classifiers trained on pre-war court decisions (2008--2013) lose 27.9 percentage points when applied to full-scale invasion era decisions (2022--2026), with a pronounced forward--backward asymmetry: newer models transfer backward (+14.6~pp above forward transfer), but older models fail catastrophically on wartime legal language. For practitioners: tokenizer analysis should precede model selection, and zero-shot is a more reliable default than few-shot for morphologically rich languages. To support reproducibility and address the absence of Ukrainian from legal NLP benchmarks, we release a public dataset of 14{,}452 court decisions spanning 2008--2026, annotated with seven outcome labels across three temporal epochs that capture the impact of armed conflict on judicial proceedings.
\end{abstract}


\section{Introduction}

The rapid proliferation of large language models (LLMs) has created an implicit hierarchy among the world's languages. English, as the dominant language in pre-training corpora, benefits from well-optimized tokenizers, extensive benchmarks, and thorough evaluation. Languages with Cyrillic scripts, complex morphology, and smaller digital footprints, such as Ukrainian, face a compounding disadvantage: their words are split into more subword tokens, resulting in higher inference costs, shorter effective context windows, and potentially degraded performance \citep{petrov2024language, ahia2023all}.

This disparity is not merely academic. For practitioners building legal technology platforms that must process tens of thousands of court decisions daily, the choice of foundation model has direct consequences for operational cost, latency, and accuracy. A model that tokenizes Ukrainian text into 60\% more tokens than an alternative is, effectively, 60\% more expensive per document, before any consideration of output quality.

In this paper, we present Experiment~A of the LEX AI Test Training program: a systematic evaluation of seven foundation models on Ukrainian legal text. Our contributions are:

\begin{enumerate}[leftmargin=*]
    \item We measure \textbf{tokenizer fertility}, the ratio of subword tokens to whitespace-delimited words, for seven models on authentic Ukrainian legal documents, revealing a 1.6$\times$ spread between the most and least efficient tokenizers.
    \item We evaluate \textbf{zero-shot and few-shot performance} on three legal NLP tasks (case type classification, case outcome classification, and legal norm extraction), finding that model size is a poor predictor of performance on Ukrainian text.
    \item We document a \textbf{counterintuitive few-shot degradation effect}: for the majority of models tested, providing task demonstrations reduces rather than improves performance on case outcome classification, with one model (Qwen~3 235B) losing 26.0 percentage points.
    \item We provide a \textbf{cost--performance analysis} across all models via AWS Bedrock, offering practitioners a directly actionable comparison.
    \item We release a \textbf{public benchmark dataset} of 14,452 court decisions spanning 2008--2026 with seven outcome labels, five jurisdiction types, and three temporal epochs reflecting major geopolitical disruptions (Crimea annexation 2014, full-scale invasion 2022), contributed to the LEXTREME benchmark as the first Cyrillic-script subset.
    \item We conduct a \textbf{cross-temporal generalization experiment} using a classical baseline (TF-IDF + logistic regression) that validates the temporal epoch structure of the released dataset, quantifies a 27.9-pp forward degradation gap caused by armed conflict, and reveals a directional asymmetry in temporal transfer that has implications for legal NLP system maintenance.
\end{enumerate}

\section{Related Work}

\subsection{Tokenizer Fertility and Multilingual Fairness}

The problem of unequal tokenization across languages has received growing attention. \citet{rust2021good} demonstrated that the monolingual performance of multilingual models correlates strongly with the proportion of pre-training data in a given language, and that tokenizer fertility is a useful proxy for this representation. \citet{petrov2024language} formalized the ``language tax'' imposed by suboptimal tokenization, showing that non-Latin-script languages can require 2--15$\times$ more tokens per semantic unit than English. \citet{ahia2023all} extended this analysis to commercial APIs, demonstrating that the cost of processing equivalent content varies by an order of magnitude across languages due to tokenizer design choices.

These studies primarily examine general-domain text. Our work focuses specifically on legal Ukrainian, a register characterized by formulaic phrasing, domain-specific terminology, and extensive citation of legislative norms, all of which interact with tokenizer vocabulary in domain-specific ways.

\subsection{Ukrainian NLP}

Ukrainian language technology has developed rapidly since 2014, driven by community efforts and increasing digitization of government data. The \textit{lang-uk} project \citep{languk2018} established foundational corpora and tools, including tokenizers, POS taggers, and NER models trained on Ukrainian web text. \citet{syvokon2023uagec} introduced UA-GEC, a grammatical error correction corpus, and demonstrated that Ukrainian-specific training data substantially outperforms multilingual transfer for morphologically sensitive tasks. \citet{chaplynskyi2023ukrbruk} contributed Ukrainian Brown Corpus resources and systematic evaluations of multilingual models on Ukrainian, showing consistent underperformance compared to English on the same architectures, a finding our work extends to the legal domain.

Despite these advances, Ukrainian NLP remains underrepresented in foundation model evaluation. No published benchmark systematically compares commercial LLMs on Ukrainian domain-specific tasks, and legal Ukrainian, with its distinct register, formulaic structures, and legislative citation conventions, has received essentially no attention in the NLP literature.

\subsection{Legal NLP}

Legal NLP has matured from rule-based systems to transformer-based approaches. LEGAL-BERT \citep{chalkidis2020legal} demonstrated the value of domain-specific pre-training for English legal text. The LEXTREME benchmark \citep{niklaus2023lextreme} extended evaluation to multiple European languages, though Ukrainian was not included. Most legal NLP benchmarks focus on Western European languages and common-law jurisdictions; civil-law systems with Cyrillic scripts remain underrepresented. To address this gap, we contribute a Ukrainian court decision dataset to LEXTREME (Section~\ref{sec:public_dataset}), providing the first Cyrillic-script legal NLP subset with temporal epoch annotations that capture the impact of armed conflict on judicial proceedings.

\subsection{Multilingual LLM Evaluation}

MMLU \citep{hendrycks2021measuring} and its multilingual extensions have become standard benchmarks for LLM capability. However, these benchmarks typically cover general knowledge and may not reflect domain-specific performance. \citet{lai2023chatgpt} evaluated ChatGPT across multiple languages and tasks, finding significant performance variation by language. Our work complements these studies by providing domain-specific (legal) evaluation on a language (Ukrainian) that is typically absent from published benchmarks.

\section{Methodology}

\subsection{Evaluation Dataset}
\label{sec:dataset}

We constructed our evaluation corpus from 300 court decisions sampled from the Unified State Register of Court Decisions (EDRSR, Ukrainian: \textit{Yedynyi Derzhavnyi Reiestr Sudovykh Rishen}), the official public repository of all Ukrainian court decisions. EDRSR contains over 120 million documents spanning 2006 to the present.

Documents were stratified by jurisdictional category with equal representation:

\begin{itemize}[leftmargin=*]
    \item \textbf{Civil} (\textit{tsyvilna}): 75 decisions
    \item \textbf{Criminal} (\textit{kryminalna}): 75 decisions
    \item \textbf{Commercial} (\textit{hospodarska}): 75 decisions
    \item \textbf{Administrative} (\textit{administratyvna}): 75 decisions
\end{itemize}

All documents are authentic court decisions in Ukrainian, extracted from the production database of the LEX AI platform (legal.org.ua). Documents were truncated to 6,000 characters for tokenizer fertility measurement to ensure consistent comparison across models with varying context windows. For task evaluation, the full document text was used, up to each model's context limit.

\subsubsection{Gold Label Construction}
\label{sec:gold_labels}

Gold labels for each task were derived as follows.

\paragraph{Case type.} Labels are taken directly from the EDRSR metadata field \texttt{justice\_kind}, which is assigned by court clerks at the time of case registration. This field is authoritative and requires no additional validation. All 300 documents carry case type labels.

\paragraph{Case outcome.} Labels were extracted from the dispositive section of each decision via a rule-based regex parser using keyword patterns for each of the five outcome categories (e.g., \foreignlanguage{ukrainian}{\textit{``позов задовольнити''}} for granted, \foreignlanguage{ukrainian}{\textit{``у задоволенні відмовити''}} for denied). To validate the parser's accuracy, we employed a three-source majority vote procedure: (1)~the regex parser, (2)~Claude Sonnet~4.5 as an independent judge classifying the same dispositive text, and (3)~NVIDIA Nemotron Super~3 as a tiebreaker for disputed cases. Of 300 documents, 205 (68\%) received identical labels from the regex parser and Claude Sonnet. The remaining documents were submitted to Nemotron Super~3 as a tiebreaker: 68 were resolved by majority vote (at least two of three sources agreed on a valid outcome label), and 27 were excluded (either all three sources disagreed, or the majority outcome was ``indeterminate''). The final validated dataset comprises 273 documents ($205 + 68$) with outcome labels confirmed by at least two independent sources.

\paragraph{Norm extraction.} Reference sets were constructed by extracting legislative citations using regex patterns matching Ukrainian citation conventions (e.g., \foreignlanguage{ukrainian}{\textit{``стаття 125''}}, \foreignlanguage{ukrainian}{\textit{``ст.~43''}}). A validation study on 30 documents using Claude Sonnet~4.5 as an independent annotator found that the regex extractor achieves 91\% precision but only 55\% recall (F1 = 0.66); it captures the most prominent citations but misses approximately 45\% of norms identified by a stronger reader. Norm extraction F1 scores reported in this paper therefore measure \emph{agreement with the regex reference set}, not agreement with the full set of legal citations in each document. This means the reported F1 likely \emph{underestimates} the true extraction capability of models that identify citations beyond the regex reference set.

\subsection{Public Benchmark Dataset}
\label{sec:public_dataset}

To address the evaluation scale limitation of our 273-document corpus and the absence of Ukrainian from the LEXTREME benchmark, we release a large-scale public dataset of 14,452 court decisions extracted from EDRSR using the same section-parsing methodology described above. The dataset spans 15 years (2008--2026) and is structured into three \emph{temporal epochs} reflecting major geopolitical disruptions to the Ukrainian judicial system:

\begin{itemize}[leftmargin=*]
    \item \textbf{Pre-war (2008--2013):} 2,610 decisions. Peacetime baseline; all 832 courts operational across all oblasts and Crimea. Criminal docket dominated by property crimes and drug offenses.
    \item \textbf{Hybrid war (2014--2021):} 4,842 decisions. Following the annexation of Crimea and the onset of the Donbas conflict, approximately 40 courts in occupied territories ceased operating under Ukrainian jurisdiction. New case categories emerged: internally displaced persons' property rights, anti-terrorist operation proceedings, and amended Criminal Code articles on terrorism (art.~258) and separatism (art.~110).
    \item \textbf{Full-scale (2022--2026):} 7,000 decisions. Full-scale invasion and martial law declaration altered procedural timelines, appeal rules, and case-type distributions. Military criminal cases surged (AWOL art.~407, desertion art.~408, draft evasion art.~336), and new Criminal Code articles were introduced (collaborationism art.~111-1, aiding the aggressor state art.~111-2).
\end{itemize}

Each decision is annotated with seven outcome labels (\textit{granted}, \textit{guilty}, \textit{partial}, \textit{closed}, \textit{denied}, \textit{plea\_deal}, \textit{other}), five jurisdiction types (civil, criminal, commercial, administrative, administrative offense), and includes both the facts section (model input) and the dispositive section (for label verification). The epoch structure enables cross-temporal generalization experiments -- e.g., training on pre-war decisions and evaluating on full-scale-era cases -- a temporal robustness challenge absent from existing legal NLP benchmarks. The dataset is available at \url{https://huggingface.co/datasets/overthelex/ukrainian-court-decisions} (config: \texttt{case\_outcome\_temporal}) and has been submitted as a pull request to the LEXTREME benchmark.

\subsection{Classical Baseline for Cross-Temporal Evaluation}
\label{sec:classical_baseline}

To validate the temporal epoch structure and establish a non-neural baseline, we train TF-IDF + logistic regression classifiers on the released dataset. For each of the three temporal epochs, we sample 5{,}000 training documents and 1{,}000 test documents (non-overlapping) with the natural label distribution. We construct a shared TF-IDF vocabulary from the union of all training data (unigrams and bigrams, minimum document frequency 10, maximum 50\%), yielding 162{,}152 features. We train $L_2$-regularized multinomial logistic regression via \texttt{glmnet} \citep{friedman2010glmnet} with 5-fold cross-validation for regularization selection ($\alpha = 0.5$). Each epoch-specific classifier is then evaluated on all three epoch test sets, producing a $3 \times 3$ accuracy matrix that quantifies cross-temporal generalization.

\subsection{Models}
\label{sec:models}

We evaluated seven models from five providers, all accessed via the AWS Bedrock API. Table~\ref{tab:models} summarizes the models and their architectures.

\begin{table}[t]
\centering
\caption{Models evaluated in Experiment~A. All models were accessed via AWS Bedrock. Size denotes total parameters; for MoE models, active parameters per forward pass are noted.}
\label{tab:models}
\begin{tabular}{llll}
\toprule
\textbf{Model} & \textbf{Provider} & \textbf{Architecture} & \textbf{Region} \\
\midrule
Llama 4 Maverick   & Meta      & 400B, 17B active, MoE-128 & us-east-1 \\
Llama 3.3 70B      & Meta      & 70B dense                  & us-east-1 \\
Mistral Large 3    & Mistral AI & 675B, 41B active, MoE-128 & us-east-1 \\
Nemotron Super 3   & NVIDIA    & 120B, 12B active, Mamba-Transf.\ MoE & eu-central-1 \\
Amazon Nova Pro    & Amazon    & Undisclosed                & eu-central-1 \\
Qwen3 235B         & Qwen      & A22B active, MoE           & eu-central-1 \\
Qwen3 32B          & Qwen      & 32B dense                  & eu-central-1 \\
\bottomrule
\end{tabular}
\end{table}

The selection criteria were: (1) availability on AWS Bedrock at the time of the experiment (April--May 2026), (2) representation of diverse tokenizer families (Llama/SentencePiece, Mistral/SentencePiece, Qwen/tiktoken-derived, Nova/proprietary), and (3) coverage of both dense and mixture-of-experts architectures.

\subsection{Tasks}
\label{sec:tasks}

We define three evaluation tasks of increasing difficulty:

\paragraph{Task 1: Case Type Classification (4-class).}
Given the full text of a court decision, classify it into one of four jurisdictional categories: civil (\textit{tsyvilna}), criminal (\textit{kryminalna}), commercial (\textit{hospodarska}), or administrative (\textit{administratyvna}). This task tests basic document understanding, as case type is typically inferable from procedural language and cited legislation.

\paragraph{Task 2: Case Outcome Classification (5-class).}
Given the full text, classify the case outcome into one of five categories: granted (\textit{zadovoleno}), denied (\textit{vidmovleno}), left without consideration (\textit{zalysheno bez rozghliadu}), partially granted (\textit{chastkovo zadovoleno}), or closed (\textit{zakryto}). This task requires understanding the dispositive section of the decision and is complicated by a severely imbalanced label distribution (see Section~\ref{sec:results_outcome}).

\paragraph{Task 3: Legal Norm Extraction (F1).}
Given the full text, extract all legal norms (law + article pairs) cited in the decision. The model must return structured JSON output with the law name and article number for each citation. We compute set-based F1 between predicted article numbers and a regex-extracted reference set. As detailed in Section~\ref{sec:gold_labels}, this reference set has high precision (91\%) but incomplete recall (55\%), so the reported F1 measures agreement with a conservative baseline rather than true extraction performance.

\subsection{Evaluation Protocol}
\label{sec:protocol}

All evaluations were conducted via the AWS Bedrock Converse API in two modes:

\begin{itemize}[leftmargin=*]
    \item \textbf{Zero-shot}: The model receives only a task instruction and the document text.
    \item \textbf{Few-shot}: The model receives the task instruction, three labeled examples (one per minority class where applicable), and the document text.
\end{itemize}

No fine-tuning, parameter-efficient or otherwise, was performed. This design choice reflects the practical scenario facing practitioners who must select a foundation model for deployment without the resources or data for domain adaptation.

For case type classification, accuracy is computed on all 300 documents (metadata labels are authoritative; see Section~\ref{sec:gold_labels}). For case outcome classification, accuracy is reported on the 273-document validated subset after excluding 27 documents with unresolved label disagreements. For norm extraction, we report the mean document-level F1 score across all 300 documents.

The temperature was set to 0 for all inference calls to ensure deterministic outputs. All metrics are reported on the 273-document validated subset for consistency across tasks. Case type metadata labels remain authoritative on the full 300-document set, but we restrict reporting to the validated subset to enable direct comparison with case outcome results.

\section{Results}

\subsection{Tokenizer Fertility}
\label{sec:results_fertility}

Table~\ref{tab:fertility} presents tokenizer fertility measurements across all seven models, computed on 100 document samples (6,000 characters each) from the evaluation corpus.

\begin{table}[t]
\centering
\caption{Tokenizer fertility on Ukrainian legal text. Fertility = average tokens per whitespace-delimited word. Lower is more efficient. Models are sorted by ascending fertility.}
\label{tab:fertility}
\begin{tabular}{lR{1.5cm}R{1.5cm}R{1.2cm}R{1.2cm}}
\toprule
\textbf{Model} & \textbf{Fert.} ($\downarrow$) & \textbf{Ch/Tok} ($\uparrow$) & \textbf{Std} & \textbf{Med.} \\
\midrule
Llama 4 Maverick  & 2.434 & 3.090 & 0.398 & 2.350 \\
Llama 3.3 70B     & 2.652 & 2.840 & 0.452 & 2.545 \\
Mistral Large 3   & 3.057 & 2.452 & 0.444 & 2.978 \\
Nemotron Super 3  & 3.082 & 2.433 & 0.453 & 3.002 \\
Nova Pro          & 3.605 & 2.069 & 0.419 & 3.515 \\
Qwen3 235B        & 3.894 & 1.917 & 0.467 & 3.794 \\
Qwen3 32B         & 3.902 & 1.913 & 0.469 & 3.804 \\
\bottomrule
\end{tabular}
\end{table}

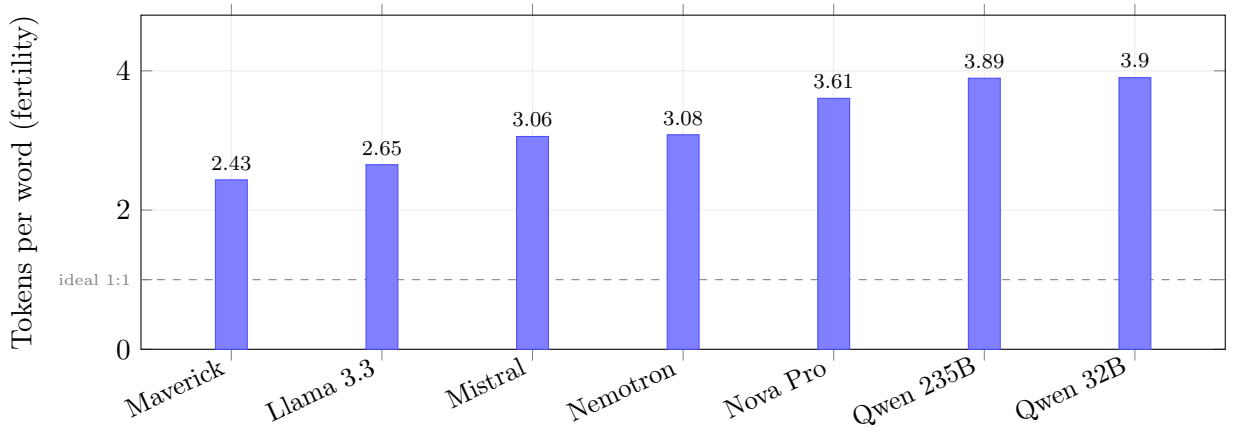
\begin{figure}[t]
\centering
\begin{tikzpicture}
\begin{axis}[
    width=\columnwidth, height=6cm,
    ybar,
    bar width=12pt,
    xlabel={},
    ylabel={Tokens per word (fertility)},
    symbolic x coords={Maverick, Llama 3.3, Mistral, Nemotron, Nova Pro, Qwen 235B, Qwen 32B},
    xtick=data,
    x tick label style={rotate=25, anchor=east, font=\small},
    ymin=0, ymax=4.8,
    grid=major,
    grid style={gray!15},
    nodes near coords,
    nodes near coords style={font=\scriptsize, above},
    every node near coord/.append style={/pgf/number format/.cd, fixed, precision=2},
    extra y ticks={1.0},
    extra y tick labels={ideal 1:1},
    extra y tick style={grid style={gray, dashed}, tick label style={font=\tiny, gray}},
]
\addplot[fill=blue!50, draw=blue!70] coordinates {
    (Maverick, 2.434) (Llama 3.3, 2.652) (Mistral, 3.057) (Nemotron, 3.082)
    (Nova Pro, 3.605) (Qwen 235B, 3.894) (Qwen 32B, 3.902)
};
\end{axis}
\end{tikzpicture}
\caption{Tokenizer fertility (average tokens per whitespace-delimited word) on 100 Ukrainian legal documents. Lower is more efficient. Llama~4 Maverick produces 38\% fewer tokens than Qwen~3 on identical text (2.43 vs.\ 3.90 tokens/word); equivalently, Qwen~3 consumes 60\% more tokens than Maverick.}
\label{fig:fertility}
\end{figure}


The results reveal a clear clustering pattern. The Llama-family tokenizers (Llama~4 Maverick and Llama~3.3) form the most efficient cluster, with fertility values of 2.43 and 2.65 tokens per word, respectively. Mistral Large~3 and Nemotron Super~3 occupy an intermediate position at approximately 3.06--3.08. The Qwen tokenizer is notably less efficient on Ukrainian text, with both Qwen~3 variants producing approximately 3.90 tokens per word, 60.3\% higher than Llama~4 Maverick.

This efficiency gap has a direct cost implication. For a typical Ukrainian court decision of 1,000 words, the Llama~4 tokenizer produces approximately 2,434 tokens, while the Qwen~3 tokenizer produces approximately 3,902, a difference of 1,468 tokens per document. At scale, this translates to substantially higher API costs for input token processing.

Notably, the two Qwen~3 models (235B and 32B) share nearly identical fertility (3.894 vs.\ 3.902), confirming that they use the same underlying tokenizer vocabulary. The same pattern holds for the Llama models, where Maverick's improved tokenizer shows an 8.2\% efficiency gain over the Llama~3.3 vocabulary.

The standard deviation of fertility is relatively consistent across models (0.398--0.469), suggesting that the efficiency differences are systematic rather than driven by outlier documents.

\subsection{Case Type Classification}
\label{sec:results_casetype}

Table~\ref{tab:case_type} presents case type classification accuracy for all models in both zero-shot and few-shot modes.

\begin{table}[t]
\centering
\caption{Case type classification accuracy (\%) on the 273-document validated subset (metadata labels are authoritative; the validated subset is used for consistency with case outcome reporting). 95\% Wilson CIs for zero-shot. Bold indicates best per mode.}
\label{tab:case_type}
\begin{tabular}{lR{1.6cm}R{2.6cm}R{1.6cm}R{1.6cm}}
\toprule
\textbf{Model} & \textbf{ZS} & \textbf{95\% CI} & \textbf{FS} & \textbf{$\Delta$} \\
\midrule
Llama 4 Maverick  & \textbf{98.9} & [96.8, 99.6] & 92.7 & $-$6.2 \\
Nemotron Super 3  & \textbf{98.9} & [96.8, 99.6] & 94.5 & $-$4.4 \\
Nova Pro          & 98.2          & [95.8, 99.2] & 92.3 & $-$5.9 \\
Qwen3 235B        & 97.4          & [94.8, 98.8] & \textbf{98.5} & $+$1.1 \\
Mistral Large 3   & 95.6          & [92.5, 97.5] & 94.9 & $-$0.7 \\
Qwen3 32B         & 95.2          & [92.0, 97.2] & 95.2 & $\pm$0.0 \\
Llama 3.3 70B     & 94.5          & [91.1, 96.6] & 96.0 & $+$1.5 \\
\bottomrule
\end{tabular}
\end{table}

Case type classification proves to be a relatively easy task, with all models achieving $\geq$92\% accuracy in at least one mode. Llama~4 Maverick and Nemotron Super~3 tie for the best zero-shot accuracy at 98.9\% (95\% CI: [96.8, 99.6]), misclassifying only 3 of 273 documents each. This advantage over Llama~3.3 70B (94.5\%) is statistically significant (McNemar $p < 0.001$), while differences among the top-4 models are not ($p > 0.05$).

A notable finding is that few-shot prompting \emph{reduces} accuracy for 4 of 7 models on this task, with the largest degradation observed for Llama~4 Maverick ($-$6.2 percentage points). This suggests that few-shot examples may confuse the model or bias it toward patterns present in the examples rather than leveraging its general understanding of Ukrainian legal document structure.

\subsection{Case Outcome Classification}
\label{sec:results_outcome}

Case outcome classification presents a substantially harder challenge. Results are reported on the 273-document validated subset (see Section~\ref{sec:gold_labels}). The label distribution is imbalanced: 230 of 273 documents (84.2\%) have the outcome ``granted'' (\textit{zadovoleno}), followed by ``left without consideration'' (21), ``denied'' (15), and ``closed'' (7). The ``partially granted'' class was entirely excluded during label validation, as all instances were disputed by the independent judge.

\begin{table}[t]
\centering
\caption{Case outcome classification accuracy (\%) on the 273-document validated subset. 95\% Wilson confidence intervals shown for zero-shot. Bold indicates best per mode.}
\label{tab:case_outcome}
\begin{tabular}{lR{1.6cm}R{2.6cm}R{1.6cm}R{1.6cm}}
\toprule
\textbf{Model} & \textbf{ZS} & \textbf{95\% CI} & \textbf{FS} & \textbf{$\Delta$} \\
\midrule
Nemotron Super 3  & \textbf{96.0} & [92.9, 97.7] & 83.2 & $-$12.8 \\
Qwen3 235B        & 93.8          & [90.3, 96.1] & 67.8 & $-$26.0 \\
Nova Pro          & 92.3          & [88.5, 94.9] & \textbf{92.7} & $+$0.4 \\
Mistral Large 3   & 91.6          & [87.7, 94.3] & 85.3 & $-$6.2 \\
Llama 4 Maverick  & 91.2          & [87.3, 94.0] & 91.9 & $+$0.7 \\
Llama 3.3 70B     & 89.7          & [85.6, 92.8] & 81.0 & $-$8.8 \\
Qwen3 32B         & 86.8          & [82.3, 90.3] & 89.7 & $+$2.9 \\
\bottomrule
\end{tabular}
\end{table}


Original scores on the full 300-document set were 10--17 percentage points lower, indicating that approximately 9\% of regex-extracted outcome labels were incorrect, primarily procedural orders misclassified as substantive decisions.

Nemotron Super~3 achieves the highest zero-shot accuracy at 96.0\% (95\% CI: [92.9, 97.7]), followed by Qwen~3 235B at 93.8\% [90.3, 96.1]. While Nemotron's advantage over Qwen~3 235B is not statistically significant by McNemar's test ($p = 0.26$), Nemotron significantly outperforms Llama~3.3 70B ($p = 0.002$), Qwen~3 32B ($p < 0.001$), Nova Pro ($p = 0.02$), and Mistral Large~3 ($p = 0.02$).

However, the most striking result is the catastrophic few-shot degradation observed for several models. Qwen~3 235B drops from 93.8\% to 67.8\% ($-$26.0 pp), and Nemotron Super~3 drops from 96.0\% to 83.2\% ($-$12.8 pp).

Analysis of per-class accuracy (Table~\ref{tab:per_class_outcome}) reveals performance variation across outcome categories. The ``partially granted'' class, which had 10 instances in the original 300-document set, was entirely removed during label validation, as all 10 instances were disputed by the independent judge. This left four outcome classes in the validated subset.

\begin{table}[t]
\centering
\caption{Per-class zero-shot accuracy (\%) for case outcome classification on the 273-document validated subset. The ``partially granted'' class was excluded during validation (all instances disputed).}
\label{tab:per_class_outcome}
\small
\begin{tabular}{lR{1.2cm}R{1.2cm}R{1.2cm}R{1.2cm}}
\toprule
\textbf{Model} & \textbf{Grant.} & \textbf{Denied} & \textbf{Left w/o} & \textbf{Closed} \\
 & \textit{n=230} & \textit{n=15} & \textit{n=21} & \textit{n=7} \\
\midrule
Nemotron   & 97.4 & 86.7 & 100.0 & 57.1 \\
Qwen3 235B & 94.3 & 93.3 & 90.5 & 85.7 \\
Nova Pro   & 92.2 & 93.3 & 100.0 & 71.4 \\
Maverick   & 91.7 & 100.0 & 85.7 & 71.4 \\
Mistral    & 91.7 & 93.3 & 85.7 & 100.0 \\
Llama 3.3  & 90.4 & 66.7 & 95.2 & 100.0 \\
Qwen3 32B  & 85.2 & 100.0 & 95.2 & 85.7 \\
\bottomrule
\end{tabular}
\end{table}

\subsubsection{Tiebreaker Bias Check}
\label{sec:tiebreaker_bias}

Because Nemotron served as one of three sources in our label validation majority vote (Section~\ref{sec:gold_labels}), its use as both tiebreaker and evaluated model could introduce systematic bias. To assess this, we partition the validated subset into \emph{easy} documents ($n{=}205$), where the regex parser and Claude Sonnet agreed without tiebreaker intervention, and \emph{hard} documents ($n{=}68$), where Nemotron's vote resolved the dispute ($205 + 68 = 273$). On the easy subset, where Nemotron had no influence on label assignment, Nemotron achieves 98.0\% (201/205), tied with Llama~4 Maverick (98.0\%) and above all other models (Qwen~3 235B 97.6\%, Llama~3.3 96.6\%, Mistral 96.1\%, Qwen~3 32B 94.6\%). Since the easy subset is free of tiebreaker influence and already shows Nemotron tied for first, the overall lead does not depend on the hard subset. On the hard subset ($n{=}68$), Nemotron achieves 89.7\% (61/68), but we cannot fully disentangle this from tiebreaker advantage; Nemotron's vote partly determined which labels were ``correct'' for these documents. We therefore base our primary ranking claims on the easy subset and the full validated set, acknowledging that hard-subset performance may be inflated for Nemotron relative to other models.

\subsection{Legal Norm Extraction}
\label{sec:results_norms}

Norm extraction requires the model to identify and structure all legal citations in a court decision, a task that combines information extraction with domain knowledge of Ukrainian legislative naming conventions.

\begin{table}[t]
\centering
\caption{Legal norm extraction mean F1 scores on the 273-document validated subset. Bold indicates best per mode.}
\label{tab:norm_extraction}
\begin{tabular}{lR{2.0cm}R{2.0cm}R{2.0cm}}
\toprule
\textbf{Model} & \textbf{Zero-Shot F1} & \textbf{Few-Shot F1} & \textbf{$\Delta$ (FS--ZS)} \\
\midrule
Llama 3.3 70B     & \textbf{0.604} & \textbf{0.606} & $+$0.001 \\
Nova Pro          & 0.575          & 0.570          & $-$0.005 \\
Mistral Large 3   & 0.561          & 0.560          & $-$0.002 \\
Nemotron Super 3  & 0.543          & 0.547          & $+$0.004 \\
Qwen3 32B         & 0.514          & 0.515          & $+$0.002 \\
Llama 4 Maverick  & 0.487          & 0.486          & $-$0.001 \\
Qwen3 235B        & 0.463          & 0.458          & $-$0.005 \\
\bottomrule
\end{tabular}
\end{table}


Llama~3.3 70B achieves the highest agreement with the regex reference set (F1 = 0.604--0.606 in both modes). The ranking on norm extraction differs markedly from classification tasks: Llama~3.3 70B, which ranks 7th on case type classification, is the clear leader here. This suggests that norm extraction relies on different capabilities, likely stronger pattern recognition for legal citation formats and better retention of long-range dependencies in document text.

Notably, few-shot prompting has minimal effect on norm extraction performance across all models, with deltas ranging from $-$0.005 to $+$0.004. The task's structured output format (JSON with law/article pairs) may already provide sufficient specification, making examples redundant.

\paragraph{Interpreting norm extraction scores.} As noted in Section~\ref{sec:gold_labels}, the regex reference set has high precision (91\%) but only 55\% recall compared to Claude Sonnet~4.5 as an independent annotator. The reported F1 scores therefore represent a \emph{lower bound} on model capability: models that correctly identify citations beyond the regex reference set are penalized as false positives. This affects all models equally and preserves the relative ranking, but means that the absolute F1 values (0.46--0.60) understate the true extraction quality. We estimate that true F1 against a comprehensive gold standard would be approximately 10--15 points higher, based on the 45\% recall gap in the reference set.

\subsection{The Few-Shot Degradation Effect}
\label{sec:fewshot}

One of the most striking findings across our experiments is the systematic degradation of performance under few-shot prompting, particularly for case outcome classification. Table~\ref{tab:fewshot_delta} summarizes the few-shot effect across all model--task combinations.

\begin{table}[t]
\centering
\caption{Few-shot effect (few-shot minus zero-shot, in percentage points). Negative values (degradation) highlighted in \textcolor{red}{red}. Improvements in \textcolor{blue}{blue}. $\Delta$ values computed from raw accuracy scores prior to rounding; minor discrepancies with tabulated rounded values may appear (e.g., for Mistral on Case Outcome, $91.6 - 85.3 = 6.3$ vs.\ reported $-6.2$).}
\label{tab:fewshot_delta}
\begin{tabular}{lR{2.0cm}R{2.0cm}R{2.0cm}}
\toprule
\textbf{Model} & \textbf{Case Type} & \textbf{Case Outc.} & \textbf{Norm Ext.} \\
\midrule
Llama 3.3 70B     & \textcolor{blue}{$+$1.5}  & \textcolor{red}{$-$8.8}  & \textcolor{blue}{$+$0.1} \\
Llama 4 Maverick  & \textcolor{red}{$-$6.2}   & \textcolor{blue}{$+$0.7} & \textcolor{red}{$-$0.1} \\
Mistral Large 3   & \textcolor{red}{$-$0.7}   & \textcolor{red}{$-$6.2}  & \textcolor{red}{$-$0.2} \\
Nemotron Super 3  & \textcolor{red}{$-$4.4}   & \textcolor{red}{$-$12.8} & \textcolor{blue}{$+$0.4} \\
Nova Pro          & \textcolor{red}{$-$5.9}   & \textcolor{blue}{$+$0.4}  & \textcolor{red}{$-$0.5} \\
Qwen3 235B        & \textcolor{blue}{$+$1.1}  & \textcolor{red}{$-$26.0} & \textcolor{red}{$-$0.5} \\
Qwen3 32B         & $\pm$0.0                  & \textcolor{blue}{$+$2.9} & \textcolor{blue}{$+$0.2} \\
\bottomrule
\end{tabular}
\end{table}

Of the 21 model--task combinations, 12 show degradation under few-shot prompting. The effect is particularly severe for case outcome classification, where 4 of 7 models perform worse with examples. The largest degradation (Qwen~3 235B, $-$26.0 pp) suggests that few-shot examples for this imbalanced task may anchor the model's predictions toward the demonstrated classes in a way that conflicts with its zero-shot prior.

We hypothesize several mechanisms:
\begin{enumerate}[leftmargin=*]
    \item \textbf{Distribution mismatch}: Few-shot examples drawn from minority classes may distort the model's prior over class frequencies.
    \item \textbf{Surface-level pattern matching}: Models may latch onto superficial features of few-shot examples (e.g., specific legal phrases) rather than learning the underlying classification rule.
    \item \textbf{Morphological interference}: Ukrainian's rich morphology means that semantically equivalent expressions have many surface forms; few-shot examples may inadvertently narrow the model's pattern space.
\end{enumerate}

\subsubsection{Stratified Few-Shot Ablation}
\label{sec:stratified_ablation}

To disentangle hypothesis~1 (distribution mismatch) from hypotheses~2--3, we conducted a stratified few-shot ablation on the two models with the largest degradation: Nemotron Super~3 and Qwen~3 235B. Instead of one example per minority class, we provided five examples matching the natural class distribution (4~granted, 1~denied), reflecting the 84\%/16\% split in the validated dataset.

\begin{table}[t]
\centering
\caption{Stratified few-shot ablation on case outcome classification (273-document validated subset). ``Minority FS'' uses one example per class; ``Stratified FS'' uses 5 examples in natural proportions (4~granted, 1~denied).}
\label{tab:stratified_ablation}
\begin{tabular}{lR{1.8cm}R{1.8cm}R{2.0cm}}
\toprule
\textbf{Model} & \textbf{Zero-Shot} & \textbf{Minority FS} & \textbf{Stratified FS} \\
\midrule
Nemotron Super 3 & \textbf{96.0} & 83.2 ($-$12.8) & 80.2 ($-$15.8) \\
Qwen3 235B       & \textbf{93.8} & 67.8 ($-$26.0) & 67.4 ($-$26.4) \\
\bottomrule
\end{tabular}
\end{table}

As Table~\ref{tab:stratified_ablation} shows, stratified few-shot examples produce degradation equal to or \emph{worse} than minority-balanced examples ($-$15.8~pp vs.\ $-$12.8~pp for Nemotron; $-$26.4~pp vs.\ $-$26.0~pp for Qwen~3 235B). This result effectively rules out distribution mismatch (hypothesis~1) as the primary cause.

\subsubsection{Prompt Sensitivity Ablation}
\label{sec:prompt_sensitivity}

To rule out prompt-specific artifacts, we tested three prompt formulations for Qwen~3 235B few-shot case outcome classification: (1)~the original Ukrainian prompt, (2)~English-language instructions with Ukrainian class labels, and (3)~a verbose Ukrainian prompt with numbered options.

\begin{table}[t]
\centering
\caption{Prompt sensitivity ablation for Qwen~3 235B on case outcome classification (few-shot, $n{=}273$). The degradation is robust across all prompt formulations.}
\label{tab:prompt_sensitivity}
\begin{tabular}{lR{2.0cm}R{2.0cm}}
\toprule
\textbf{Prompt variant} & \textbf{Accuracy} & \textbf{$\Delta$ vs.\ ZS} \\
\midrule
Zero-shot (baseline)          & 93.8 & --- \\
\midrule
Ukrainian instructions (orig) & 49.1 & $-$44.7 \\
English instructions           & 60.1 & $-$33.7 \\
Verbose Ukrainian              & 60.1 & $-$33.7 \\
\bottomrule
\end{tabular}
\end{table}

As Table~\ref{tab:prompt_sensitivity} shows, the few-shot degradation is robust across all three prompt formulations, with accuracy dropping by 34--45 percentage points regardless of instruction language or verbosity. English-language instructions partially mitigate the effect ($-$33.7~pp vs.\ $-$44.7~pp), suggesting that the interference operates partly at the level of Ukrainian-language demonstration parsing. However, even with English instructions, few-shot performance (60.1\%) remains far below zero-shot (93.8\%), confirming that the degradation is not an artifact of a single prompt template. The combined evidence from stratified example selection (Section~\ref{sec:stratified_ablation}) and prompt variation rules out both distribution mismatch and prompt-specific confounds, supporting the morphological interference hypothesis.

\begin{figure}[t]
\centering
\begin{tikzpicture}
\begin{axis}[
    width=\columnwidth, height=7.5cm,
    ybar,
    bar width=3.5pt,
    xlabel={Model},
    ylabel={Few-shot $-$ Zero-shot (pp)},
    symbolic x coords={Maverick, Llama 3.3, Mistral, Nemotron, Nova Pro, Qwen 235B, Qwen 32B},
    xtick=data,
    x tick label style={rotate=25, anchor=east, font=\small},
    ymin=-30, ymax=6,
    grid=major,
    grid style={gray!15},
    legend style={at={(0.02,0.02)}, anchor=south west, font=\small},
    every axis plot/.append style={fill opacity=0.8},
    extra y ticks={0},
    extra y tick style={grid style={black, thick}},
]
\addplot[fill=blue!60, draw=blue!80] coordinates {
    (Maverick, -6.2) (Llama 3.3, 1.5) (Mistral, -0.7) (Nemotron, -4.4)
    (Nova Pro, -5.9) (Qwen 235B, 1.1) (Qwen 32B, 0.0)
};
\addplot[fill=red!60, draw=red!80] coordinates {
    (Maverick, 0.7) (Llama 3.3, -8.8) (Mistral, -6.2) (Nemotron, -12.8)
    (Nova Pro, 0.4) (Qwen 235B, -26.0) (Qwen 32B, 2.9)
};
\addplot[fill=green!50!black, draw=green!60!black] coordinates {
    (Maverick, -0.1) (Llama 3.3, 0.1) (Mistral, -0.2) (Nemotron, 0.4)
    (Nova Pro, -0.5) (Qwen 235B, -0.5) (Qwen 32B, 0.2)
};
\legend{Case Type, Case Outcome, Norm Extraction}
\end{axis}
\end{tikzpicture}
\caption{Few-shot effect (few-shot minus zero-shot, in percentage points) across all model--task combinations. Bars below the zero line indicate degradation. Case outcome classification (red) shows the most severe and widespread degradation, with Qwen~3 235B dropping 26~pp. Norm extraction (green) is largely unaffected by few-shot prompting.}
\label{fig:fewshot_delta}
\end{figure}
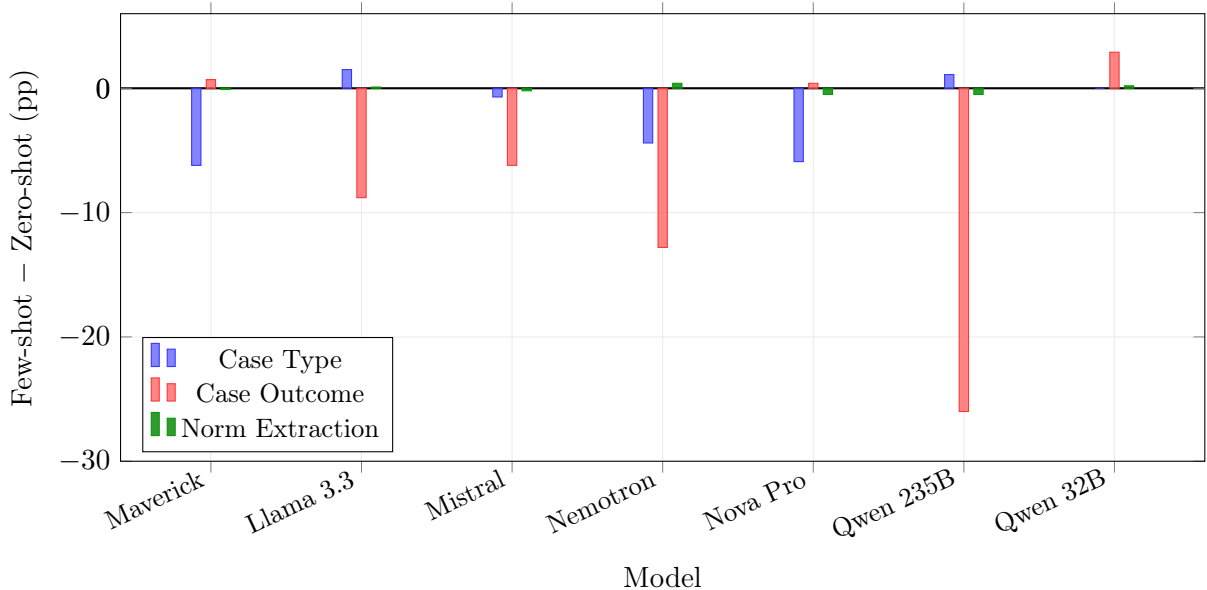

Figure~\ref{fig:fewshot_delta} visualizes the few-shot effect across all model--task combinations.

\subsection{Composite Ranking}
\label{sec:composite}

To provide a holistic comparison, we compute two composite scores. The \emph{3-task composite} is the unweighted mean of case type accuracy, case outcome accuracy, and norm extraction F1 (scaled to 0--100). Because the norm extraction gold standard has incomplete recall (Section~\ref{sec:gold_labels}), we also report a \emph{classification-only composite}, the mean of case type and case outcome accuracy, which relies exclusively on validated labels and is unaffected by reference set limitations. Table~\ref{tab:composite} presents both rankings.

\begin{table}[t]
\centering
\caption{Composite ranking by zero-shot performance and cost. 3-task composite $= (\text{CT} + \text{CO} + 100 \cdot \text{NE}_{\text{F1}}) / 3$; classification-only $= (\text{CT} + \text{CO})/2$. Models sorted by 3-task composite.}
\label{tab:composite}
\small
\begin{tabular}{lR{1.1cm}R{1.1cm}R{1.0cm}R{1.2cm}R{1.2cm}R{1.2cm}}
\toprule
\textbf{Model} & \textbf{CT} & \textbf{CO} & \textbf{NE} & \textbf{3-task} & \textbf{Cls-only} & \textbf{Cost} \\
 & \textbf{\%} & \textbf{\%} & \textbf{F1} & & & \textbf{(\$)} \\
\midrule
Nemotron Super 3  & 98.9 & 96.0 & .543 & \textbf{83.1} & \textbf{97.5} & 3.61 \\
Nova Pro          & 98.2 & 92.3 & .575 & 82.7 & 95.2 & 4.98 \\
Llama 3.3 70B     & 94.5 & 89.7 & .604 & 81.6 & 92.1 & 3.00 \\
Mistral Large 3   & 95.6 & 91.6 & .561 & 81.1 & 93.6 & 10.99 \\
Llama 4 Maverick  & 98.9 & 91.2 & .487 & 79.6 & 95.1 & 0.81 \\
Qwen3 235B        & 97.4 & 93.8 & .463 & 79.2 & 95.6 & 5.37 \\
Qwen3 32B         & 95.2 & 86.8 & .514 & 77.8 & 91.0 & 2.64 \\
\bottomrule
\end{tabular}
\end{table}

Nemotron Super~3 ranks first under \emph{both} composite metrics (83.1 and 97.5), confirming that its lead is robust to the choice of aggregation. The classification-only composite, which avoids the regex reference set limitation, shows a tighter field: Nemotron (97.5), Qwen~3 235B (95.6), Nova Pro (95.2), and Maverick (95.1) are separated by only 2.4 points. This highlights that the 3-task composite's wider spread is partly driven by norm extraction score differences, which, as discussed in Section~\ref{sec:results_norms}, underestimate the true capability of models that identify citations beyond the regex reference set.

On the cost dimension, Llama~4 Maverick costs only \$0.81 for the entire experiment, while Mistral Large~3 costs \$10.99, a 13.6$\times$ cost difference. Under the classification-only composite, Maverick (95.1 at \$0.81) achieves 97\% of Nemotron's quality (97.5 at \$3.61) at 22\% of the cost.

\begin{figure}[t]
\centering
\begin{tikzpicture}
\begin{axis}[
    width=\columnwidth, height=7cm,
    xlabel={Total experiment cost (USD)},
    ylabel={Composite score},
    xmin=0, xmax=12,
    ymin=76, ymax=84.5,
    grid=major,
    grid style={gray!20},
    legend pos=south east,
    legend style={font=\small},
    every node near coord/.append style={font=\scriptsize, anchor=south west},
]
\addplot[only marks, mark=*, mark size=4pt, blue!80, fill=blue!40]
    coordinates {(0.81, 79.6)} node[above right, font=\scriptsize] {Maverick};
\addplot[only marks, mark=*, mark size=4pt, red!80, fill=red!40]
    coordinates {(3.00, 81.6)} node[above right, font=\scriptsize] {Llama 3.3};
\addplot[only marks, mark=*, mark size=4pt, purple!80, fill=purple!40]
    coordinates {(3.61, 83.1)} node[above left, font=\scriptsize] {\textbf{Nemotron}};
\addplot[only marks, mark=*, mark size=4pt, cyan!80, fill=cyan!40]
    coordinates {(2.64, 77.8)} node[below right, font=\scriptsize] {Qwen3 32B};
\addplot[only marks, mark=*, mark size=4pt, pink!80, fill=pink!40]
    coordinates {(4.98, 82.7)} node[above right, font=\scriptsize] {Nova Pro};
\addplot[only marks, mark=*, mark size=4pt, violet!80, fill=violet!40]
    coordinates {(5.37, 79.2)} node[below right, font=\scriptsize] {Qwen3 235B};
\addplot[only marks, mark=*, mark size=4pt, green!60!black, fill=green!30]
    coordinates {(10.99, 81.1)} node[above left, font=\scriptsize] {Mistral};
\addplot[dashed, gray, thick] coordinates {(0.81, 79.6) (3.00, 81.6) (3.61, 83.1)};
\end{axis}
\end{tikzpicture}
\caption{Cost--quality frontier for seven models on Ukrainian legal text. Each point represents one model; the dashed line traces the Pareto frontier. Nemotron Super~3 offers the best composite score at moderate cost; Maverick occupies the efficient corner. Mistral Large~3, despite 5.6$\times$ more total parameters (3.4$\times$ active), delivers lower quality at 3$\times$ the cost of Nemotron.}
\label{fig:cost_composite}
\end{figure}
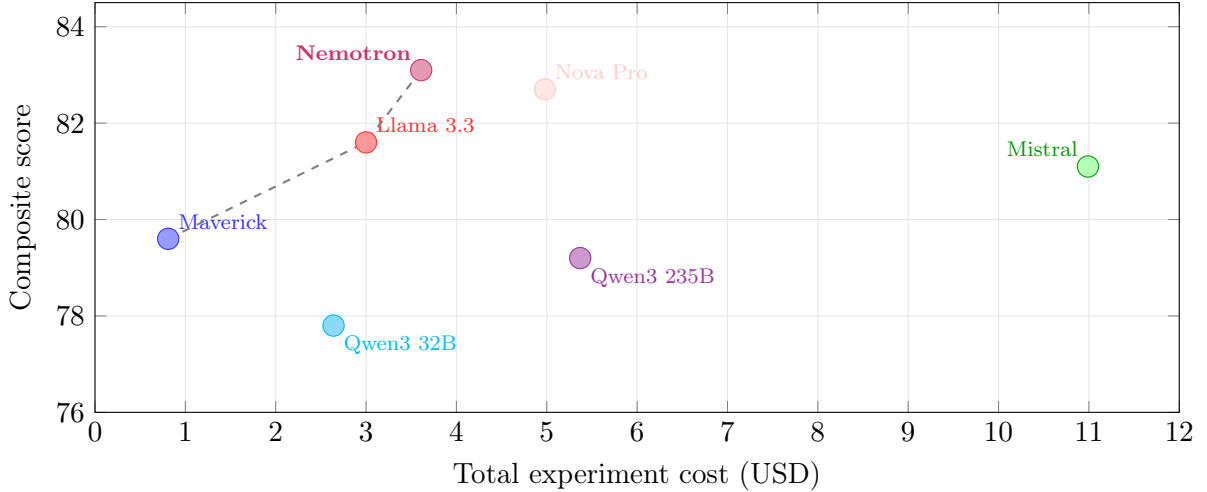

Figure~\ref{fig:cost_composite} visualizes the cost--quality frontier across all seven models.

\subsection{Cost Analysis}
\label{sec:cost}

Table~\ref{tab:costs} presents detailed cost breakdowns by model. Costs reflect actual API charges via AWS Bedrock during the experiment period.

\begin{table}[t]
\centering
\caption{Total experiment cost per model (USD), covering all tasks in both zero-shot and few-shot modes ($\approx$1,800 inference calls per model). Sorted by ascending cost.}
\label{tab:costs}
\begin{tabular}{lR{2.0cm}R{2.0cm}}
\toprule
\textbf{Model} & \textbf{Total Cost} & \textbf{Cost/Call} \\
\midrule
Llama 4 Maverick  & \$0.81  & \$0.00045 \\
Qwen3 32B         & \$2.64  & \$0.00147 \\
Llama 3.3 70B     & \$3.00  & \$0.00167 \\
Nemotron Super 3  & \$3.61  & \$0.00201 \\
Nova Pro          & \$4.98  & \$0.00277 \\
Qwen3 235B        & \$5.37  & \$0.00298 \\
Mistral Large 3   & \$10.99 & \$0.00611 \\
\midrule
\textbf{Total}    & \textbf{\$31.41} & --- \\
\bottomrule
\end{tabular}
\end{table}

The cost variation is dramatic. Llama~4 Maverick is 13.6$\times$ cheaper than Mistral Large~3 per inference call. This cost advantage derives from two factors: (1) Maverick's superior tokenizer fertility reduces input token count by 8--38\% relative to other models, and (2) Maverick's per-token pricing on Bedrock is among the lowest in the evaluated set.

Crucially, this cost advantage does not come at the expense of quality. Maverick achieves the best or tied-best zero-shot accuracy on case type classification (98.9\%) and competitive performance on case outcome classification (91.2\%, 4th place). Its relative weakness is norm extraction (F1 = 0.487, 6th place), suggesting that the smaller active parameter count (17B) may limit performance on complex extraction tasks.

\paragraph{Beyond API pricing: deployment flexibility.}
Our cost analysis reflects managed API pricing on AWS Bedrock, which is the most accessible deployment mode but not the only one. A critical distinction among our evaluated models is \emph{self-hosting capability}. Nemotron Super~3, Llama~3.3, and Llama~4 Maverick are open-weight models that can be deployed on-premises or in private clouds: Nemotron via NVIDIA NIM (NVIDIA Inference Microservices), Llama models via vLLM, TGI, or similar serving stacks. This enables organizations with GPU infrastructure to eliminate per-token API costs entirely, paying only for compute. For a legal technology platform processing millions of court decisions, the total cost of ownership (TCO) under self-hosted deployment can be an order of magnitude lower than managed API pricing.

In contrast, Amazon Nova Pro and Mistral Large~3 are available exclusively through managed APIs (Bedrock and Mistral's platform, respectively), offering no self-hosting option. Qwen~3 models are open-weight and deployable via standard inference stacks (vLLM, SGLang, TensorRT-LLM), though without the enterprise tooling and support that NVIDIA NIM provides for Nemotron.

This deployment asymmetry further strengthens Nemotron's position: it combines the highest task accuracy in our evaluation with the flexibility to be self-hosted via NIM on NVIDIA GPUs, enabling fine-tuning, domain adaptation, and data-sovereign deployment, all critical requirements for legal technology platforms handling sensitive court documents.

\subsection{Cross-Temporal Generalization}
\label{sec:temporal}

To validate the temporal epoch structure of the released dataset and establish a classical baseline, we train TF-IDF + logistic regression classifiers on each epoch's data and evaluate on all three epochs (Section~\ref{sec:classical_baseline}). Table~\ref{tab:temporal_matrix} presents the $3 \times 3$ accuracy matrix.

\begin{table}[t]
\centering
\caption{Cross-temporal generalization matrix (accuracy \%). Rows: training epoch; columns: test epoch. Diagonal entries (bold) are in-epoch baselines. Classical baseline: TF-IDF + $L_2$-regularized logistic regression, 5{,}000 train / 1{,}000 test per epoch.}
\label{tab:temporal_matrix}
\begin{tabular}{lR{1.8cm}R{1.8cm}R{2.0cm}}
\toprule
\textbf{Train $\backslash$ Test} & \textbf{Pre-war} & \textbf{Hybrid} & \textbf{Full-scale} \\
\midrule
Pre-war (2008--13)     & \textbf{86.5} & 76.3          & 58.6 \\
Hybrid (2014--21)      & 83.2          & \textbf{83.7} & 64.5 \\
Full-scale (2022--26)  & 79.5          & 80.4          & \textbf{69.3} \\
\bottomrule
\end{tabular}
\end{table}

Three patterns emerge from the cross-temporal evaluation:

\paragraph{Forward transfer fails catastrophically.}
A classifier trained on pre-war decisions loses 27.9 percentage points when applied to full-scale era cases (86.5\% $\rightarrow$ 58.6\%). The hybrid$\rightarrow$full-scale gap is smaller but still substantial ($-$19.2~pp). This degradation reflects genuine distributional shift: the full-scale invasion introduced new Criminal Code articles (collaborationism art.~111-1, aiding the aggressor art.~111-2), martial-law procedural modifications, and a surge in military criminal cases (AWOL art.~407, desertion art.~408) that have no precedent in pre-war training data.

\paragraph{Backward transfer is robust.}
The asymmetry between forward and backward transfer is striking. Models trained on newer data generalize well to older epochs: full-scale$\rightarrow$pre-war achieves 79.5\% (only $-$7.0~pp below the pre-war in-epoch baseline), while pre-war$\rightarrow$full-scale drops by $-$27.9~pp. Mean backward transfer accuracy (81.0\%) exceeds mean forward transfer (66.5\%) by 14.6~pp. This asymmetry is consistent with a \emph{legal language accumulation} hypothesis: newer legal language subsumes older conventions (courts still apply pre-war statutes), but older training data cannot anticipate future legislative and procedural changes.

\paragraph{Task difficulty increases with conflict intensity.}
In-epoch accuracy decreases monotonically across epochs: 86.5\% (pre-war) $>$ 83.7\% (hybrid) $>$ 69.3\% (full-scale). This gradient reflects increasing case-type heterogeneity and outcome unpredictability under martial law, where procedural norms are modified by presidential decrees and new offense categories produce cases without established judicial precedent.

\begin{figure}[t]
\centering
\begin{tikzpicture}
\begin{axis}[
    width=0.75\columnwidth,
    height=0.65\columnwidth,
    colormap={temporal}{
        rgb255(0cm)=(215,48,39);
        rgb255(1cm)=(254,224,139);
        rgb255(2cm)=(26,152,80)
    },
    colorbar,
    colorbar style={
        ylabel={Accuracy (\%)},
        ylabel style={font=\small},
        ytick={60,65,70,75,80,85},
    },
    point meta min=58,
    point meta max=87,
    xtick={0,1,2},
    xticklabels={Pre-war\\(08--13), Hybrid\\(14--21), Full-scale\\(22--26)},
    ytick={0,1,2},
    yticklabels={Pre-war\\(08--13), Hybrid\\(14--21), Full-scale\\(22--26)},
    x tick label style={align=center, font=\small},
    y tick label style={align=center, font=\small},
    xlabel={Test epoch},
    ylabel={Train epoch},
    xlabel style={font=\small},
    ylabel style={font=\small},
    enlargelimits=false,
    axis on top,
    nodes near coords={\pgfmathprintnumber\pgfplotspointmeta},
    every node near coord/.append style={font=\small\bfseries, black},
]
\addplot[matrix plot*, mesh/cols=3, mesh/rows=3, point meta=explicit]
    coordinates {
        (0,0) [86.5]  (1,0) [76.3]  (2,0) [58.6]
        (0,1) [83.2]  (1,1) [83.7]  (2,1) [64.5]
        (0,2) [79.5]  (1,2) [80.4]  (2,2) [69.3]
    };
\end{axis}
\end{tikzpicture}
\caption{Cross-temporal generalization heatmap. Rows: training epoch; columns: test epoch. The color gradient reveals a strong forward-degradation pattern: classifiers trained on older data (bottom row) fail on newer test data (right column), while backward transfer (top-left) is substantially more robust.}
\label{fig:temporal_heatmap}
\end{figure}
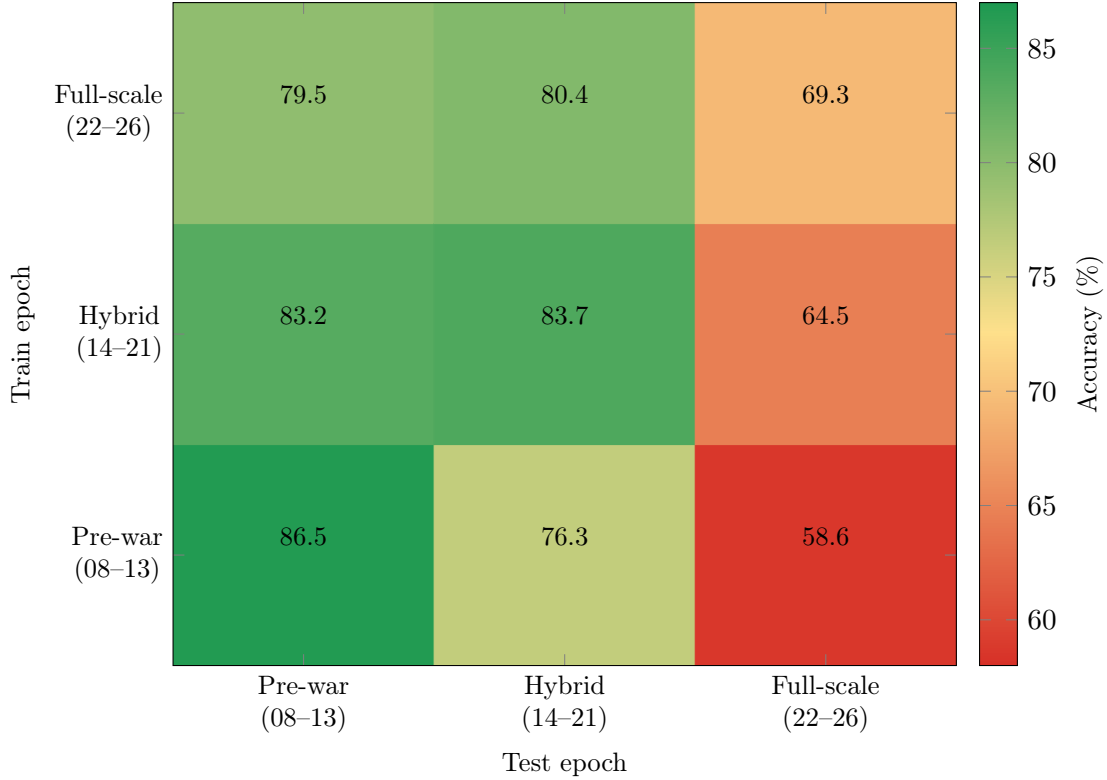

\paragraph{Comparison with foundation models.}
The zero-shot foundation models evaluated in Sections~\ref{sec:results_casetype}--\ref{sec:results_outcome} achieve 86.8--98.9\% on case type classification and 86.8--96.0\% on case outcome, both on the 273-document validated subset spanning all epochs. The TF-IDF baseline's in-epoch accuracy (69.3--86.5\% on the three-class outcome task) establishes that zero-shot foundation models substantially outperform classical methods on this task, even without fine-tuning. However, foundation models are not immune to temporal drift: our Bedrock evaluation (Section~\ref{sec:temporal_bedrock_comparison}) shows that both Nemotron Super~3 and Qwen~3 235B exhibit epoch-dependent performance, with full-scale era accuracy 25--29~pp below pre-war levels in zero-shot mode, a degradation comparable in magnitude to the classical baseline.

\label{sec:temporal_bedrock_comparison}
\paragraph{Foundation model epoch analysis.}
To compare temporal sensitivity across paradigms, we evaluated Nemotron Super~3 and Qwen~3 235B on 504 temporally stratified documents from the released dataset (168 per epoch, balanced labels). Table~\ref{tab:temporal_bedrock} presents accuracy by epoch.

\begin{table}[t]
\centering
\caption{Foundation model accuracy (\%) by temporal epoch on 504 stratified documents (168 per epoch, balanced labels across three outcome classes). Zero-shot, temperature 0.}
\label{tab:temporal_bedrock}
\begin{tabular}{lR{1.8cm}R{1.8cm}R{2.0cm}R{1.5cm}}
\toprule
\textbf{Model} & \textbf{Pre-war} & \textbf{Hybrid} & \textbf{Full-scale} & \textbf{$\Delta$} \\
\midrule
Nemotron Super 3  & 90.5 & 82.7 & 61.3 & $-$29.2 \\
Qwen3 235B        & 88.1 & 86.3 & 66.1 & $-$22.0 \\
\midrule
TF-IDF (in-epoch) & 86.5 & 83.7 & 69.3 & $-$17.2 \\
\bottomrule
\end{tabular}
\end{table}

Both foundation models show monotonically decreasing accuracy across epochs, with Nemotron dropping 29.2~pp from pre-war to full-scale and Qwen~3 dropping 22.0~pp. Notably, the TF-IDF in-epoch baseline (69.3\%) \emph{outperforms} both foundation models on full-scale era data (Nemotron: 61.3\%, Qwen~3: 66.1\%), despite the foundation models having access to vastly more parametric knowledge. This suggests that full-scale era cases present a distributional challenge that neither pre-training scale nor zero-shot prompting can overcome without epoch-specific adaptation.

\section{Discussion}

\subsection{Tokenizer Efficiency as a First-Order Concern}

Our results demonstrate that tokenizer fertility should be a first-order consideration when selecting foundation models for non-English NLP. The 1.6$\times$ fertility gap between the most and least efficient tokenizers on Ukrainian text has direct, quantifiable consequences: 60\% higher token consumption per document, 60\% higher API costs at equivalent pricing, and a proportionally reduced effective context window.

The clustering of fertility by tokenizer family, rather than model size, confirms that this is a vocabulary design choice, not an emergent property of scale. Both Qwen~3 models (32B and 235B) exhibit nearly identical fertility (3.902 vs.\ 3.894), and both Llama models cluster at the efficient end. Practitioners evaluating models for non-English deployment should therefore begin with tokenizer analysis before investing in task-specific benchmarking.

The Llama~4 tokenizer's efficiency improvement over Llama~3.3 (2.434 vs.\ 2.652, an 8.2\% reduction) indicates that Meta has actively improved Cyrillic representation between model generations, likely by expanding the vocabulary with additional Ukrainian and related-language subword units.

\subsection{Model Size Does Not Predict Ukrainian Performance}

A striking finding is the poor correlation between model size (total parameters) and Ukrainian-language task performance. Nemotron Super~3 (120B total, 12B active) achieves the highest composite score, outperforming Mistral Large~3 (675B total, 41B active) on all three tasks while costing one-third as much. Llama~4 Maverick, with only 17B active parameters, matches or exceeds 70B+ models on classification tasks.

This disconnect suggests that Ukrainian-language capability depends more on (1) the proportion and quality of Ukrainian text in pre-training data, (2) tokenizer design, and (3) instruction-following quality on non-English prompts than on raw parameter count. For practitioners, the implication is clear: model selection for low-resource languages cannot be based on English-language benchmarks alone.

\subsection{Why Nemotron Leads: Architecture and Training Hypotheses}

Nemotron Super~3's dominance on Ukrainian legal text, particularly its 96.0\% case outcome accuracy (4+ percentage points above the next-best model), warrants explanation. The model (Bedrock ID: \texttt{nvidia.nemotron-super-3-120b}, listed as ``NVIDIA Nemotron 3 Super 120B A12B'') is a 120B-parameter open model with only 12B active parameters per token, built on a \emph{hybrid Mamba-Transformer} architecture with latent mixture-of-experts (MoE). This is not a distilled Llama variant; it is a distinct architecture trained from scratch on over 10 trillion tokens, including synthetic data generated by frontier reasoning models. We hypothesize that four architectural features contribute to its performance on Ukrainian legal text.

First, \textbf{hybrid Mamba-Transformer layers}. Nemotron Super~3 combines Mamba layers (a selective state-space model offering 4$\times$ greater memory and compute efficiency than standard attention) with transformer layers for reasoning. This hybrid architecture is particularly well-suited to long legal documents: Mamba layers efficiently encode the formulaic, repetitive structure of court decisions (procedural history, cited legislation), while transformer layers handle the reasoning-intensive dispositive section. Our evaluation documents average 10,800 characters, a length where Mamba's sub-quadratic sequence scaling provides a meaningful advantage over pure transformer architectures.

Second, \textbf{latent MoE routing}. Nemotron activates only 12B of its 120B parameters per token, routing each token to four specialist experts for the computational cost of one dense forward pass. While other MoE models in our evaluation have even higher sparsity ratios (Llama~4 Maverick activates 4\% of its 400B parameters, Qwen~3 235B activates 9\%), Nemotron's \emph{latent} MoE architecture routes through four specialists per token rather than a single expert, increasing effective capacity without a proportional increase in compute cost. Combined with sub-quadratic Mamba layers, this enables Nemotron to store diverse knowledge, potentially including Ukrainian legal patterns, across 120B parameters while maintaining the inference speed of a 12B model.

Third, \textbf{synthetic training data from frontier models}. NVIDIA's training pipeline uses synthetic data generated by frontier reasoning models (likely GPT-4-class) across multiple languages. If the frontier teacher generated Ukrainian-language training samples, including legal reasoning patterns, Nemotron would inherit multilingual legal reasoning capability without requiring massive Ukrainian web corpora in the pre-training set. This synthetic data strategy may explain why Nemotron outperforms models trained primarily on organic web data, where Ukrainian is underrepresented.

Fourth, \textbf{multi-token prediction}. Nemotron employs a multi-token prediction objective during training, which has been shown to improve both inference speed and output coherence. For structured tasks such as case outcome classification, where the answer is a short Ukrainian phrase, multi-token prediction may enable more confident single-step output rather than token-by-token generation.

We note that Nemotron's tokenizer fertility (3.08 tokens/word) clusters with Mistral (3.06) rather than with the Llama family (2.43--2.65), confirming that Nemotron uses its own vocabulary rather than inheriting Llama's. Despite this moderate fertility, Nemotron's low active parameter count (12B) keeps per-token inference cost competitive: at \$0.15/M input tokens on Bedrock, it is among the cheapest models in our evaluation on a per-quality-point basis.

\subsection{The Few-Shot Paradox for Morphologically Rich Languages}

The systematic few-shot degradation we observe, particularly the 26.0-point drop for Qwen~3 235B on case outcome classification, extends a growing body of evidence on few-shot failure modes. \citet{lu2022order} showed that few-shot performance is highly sensitive to example ordering, with accuracy varying by up to 30 percentage points depending on permutation. \citet{min2022rethinking} demonstrated that few-shot demonstrations often function as format specifiers rather than task learners: ground-truth labels in examples can be replaced with random labels with minimal performance impact, suggesting that models anchor on surface-level patterns rather than learning the task. Our findings add a new dimension: for morphologically rich languages such as Ukrainian, few-shot demonstrations may actively interfere with the model's zero-shot capabilities.

For Ukrainian legal text, we hypothesize that the rich morphological system creates a combinatorial explosion of surface forms for semantically equivalent expressions. Few-shot examples, which necessarily present a tiny sample of these forms, may inadvertently narrow the model's attention to specific morphological patterns that do not generalize. In contrast, zero-shot prompting allows the model to leverage its full distributional knowledge of Ukrainian without surface-level anchoring.

Our stratified few-shot ablation (Section~\ref{sec:stratified_ablation}) provides direct evidence for this interpretation. When we replaced minority-balanced examples with examples matching the natural class distribution (4~granted, 1~denied), the degradation persisted or worsened ($-$15.8~pp for Nemotron, $-$26.4~pp for Qwen~3 235B). This rules out distribution mismatch as the primary cause and implicates the act of providing Ukrainian-language demonstrations itself as the source of interference.

This finding has practical implications: for production systems processing Ukrainian legal text, zero-shot prompting should be the default baseline, and few-shot prompting should be validated per-model and per-task rather than assumed to help.

\subsection{Task-Specific Strengths and Multi-Model Routing}
\label{sec:routing}

No single model dominates all tasks. The task-specific rankings reveal complementary strengths that motivate a routing architecture:

\begin{itemize}[leftmargin=*]
    \item \textbf{Case type classification}: Llama~4 Maverick and Nemotron Super~3 (98.9\% each). This is the easiest task, and the cheapest model (Maverick, \$0.00045/call) matches the best.
    \item \textbf{Case outcome classification}: Nemotron Super~3 (96.0\%). The hardest classification task, where Nemotron's hybrid Mamba-Transformer architecture and synthetic multilingual training data provide a clear edge.
    \item \textbf{Norm extraction}: Llama~3.3 70B (F1 = 0.604). The only model with a dense 70B architecture in our set, it excels at structured JSON extraction from long legal citations.
\end{itemize}

\paragraph{Proposed routing architecture.}
For a production legal NLP pipeline processing Ukrainian court decisions, we propose a three-tier routing strategy that assigns each document to the optimal model per task:

\begin{enumerate}[leftmargin=*]
    \item \textbf{Tier 1: Case type classification $\rightarrow$ Llama~4 Maverick.} At 98.9\% accuracy and \$0.00045/call, Maverick provides near-perfect classification at the lowest cost. Its superior tokenizer (2.43 tokens/word) further reduces input cost. This is a high-volume, low-stakes call suitable for the cheapest model.
    \item \textbf{Tier 2: Case outcome classification $\rightarrow$ Nemotron Super~3.} At 96.0\% accuracy and \$0.00201/call, Nemotron is 4.5$\times$ more expensive than Maverick per call but provides the most reliable outcome extraction, a high-stakes determination that affects downstream legal analysis.
    \item \textbf{Tier 3: Norm extraction $\rightarrow$ Llama~3.3 70B.} At F1 = 0.604 and \$0.00167/call, Llama~3.3 provides the best structured extraction. This task is typically run selectively (on documents requiring citation analysis), not on every document.
\end{enumerate}

\paragraph{Cost--quality comparison.}
Table~\ref{tab:routing} compares the proposed routing ensemble against single-model baselines for a hypothetical workload of 10,000 documents, where all documents require case type and outcome classification, and 20\% require norm extraction.

\begin{table}[t]
\centering
\caption{Cost--quality comparison for 10,000 documents: routed ensemble vs.\ single-model baselines. ``Quality'' is the 3-task composite from Table~\ref{tab:composite}: $(\text{CT} + \text{CO} + 100 \cdot \text{NE}_{\text{F1}}) / 3$. Routed cost assumes Maverick for case type (10K calls), Nemotron for case outcome (10K calls), and Llama~3.3 for norm extraction (2K calls). Single-model strategies use the same model for all tasks (22K calls: 10K CT + 10K CO + 2K NE).}
\label{tab:routing}
\small
\begin{tabular}{lR{2.0cm}R{2.0cm}R{2.0cm}}
\toprule
\textbf{Strategy} & \textbf{Cost} & \textbf{Quality} & \textbf{Cost/Quality} \\
\midrule
Routed ensemble         & \$27.94  & 85.1 & \$0.33 \\
Nemotron only           & \$44.22  & 83.1 & \$0.53 \\
Maverick only           & \$9.90   & 79.6 & \$0.12 \\
Mistral Large~3 only    & \$134.42 & 81.1 & \$1.66 \\
\bottomrule
\end{tabular}
\end{table}

The routed ensemble achieves the highest composite score (85.1) by assigning each task to the best-performing model, at a cost of \$27.94 per 10K documents. This is 37\% cheaper than using Nemotron alone (\$44.22) while delivering higher quality, because Maverick handles the easy classification tier at lower per-call cost. The Maverick-only strategy is the cheapest (\$9.90) but sacrifices 5.5 composite points, primarily on case outcome (91.2\% vs.\ 96.0\%) and norm extraction (0.487 vs.\ 0.604). Mistral Large~3, despite competitive accuracy, is 4.8$\times$ more expensive than the routed ensemble for lower quality.

This analysis assumes Bedrock API pricing. Under self-hosted deployment via NVIDIA NIM, the Nemotron and Llama tiers would have near-zero marginal cost after GPU amortization, making the routed ensemble even more cost-effective.

\subsection{Temporal Drift as a First-Class Challenge}

The cross-temporal generalization experiment (Section~\ref{sec:temporal}) reveals that temporal robustness is not merely a theoretical concern for legal NLP -- it is a dominant source of performance degradation, exceeding the impact of model selection. The 27.9-pp forward degradation gap (pre-war$\rightarrow$full-scale) is larger than the difference between the best and worst foundation models on any single task in our evaluation (12.1~pp on case outcome). This finding has two implications.

First, \textbf{legal NLP systems require temporal maintenance}. A model deployed on pre-war training data will fail on wartime cases, not because the model is weak, but because the legal landscape has shifted. The accumulation of new statutes (art.~111-1, 111-2, 407, 408), procedural modifications under martial law, and novel case-type distributions creates distributional shift that neither pre-training scale nor zero-shot prompting can absorb. Systems must be periodically re-evaluated -- and potentially re-trained -- on contemporary data.

Second, \textbf{the forward--backward asymmetry suggests that legal language is additive}. Courts continue to apply pre-war statutes alongside new wartime legislation; a model trained on 2022--2026 data has seen both old and new legal frameworks. Conversely, a 2008--2013 model has never encountered martial-law proceedings. This asymmetry distinguishes legal temporal drift from standard concept drift: the target distribution does not simply change -- it expands, accumulating new concepts without discarding old ones.

\subsection{Implications for Practitioners}

For teams building legal NLP systems for Ukrainian or other Cyrillic-script languages, we offer the following recommendations:

\begin{enumerate}[leftmargin=*]
    \item \textbf{Start with tokenizer analysis.} Before benchmarking task performance, measure tokenizer fertility on representative domain text. A 1.6$\times$ fertility difference compounds across every inference call.
    \item \textbf{Default to zero-shot.} Do not assume that few-shot prompting will help. For morphologically rich languages, validate few-shot against zero-shot per model and per task.
    \item \textbf{Ignore parameter counts.} Model size does not predict non-English performance. A 120B model outperformed a 675B model on all tasks.
    \item \textbf{Route by task, not by model.} Match model strengths to task requirements. Cheap models suffice for easy classification; invest in stronger models only for hard tasks.
    \item \textbf{Budget for temporal maintenance.} Legal NLP performance degrades with distributional shift from legislative changes. The 28-pp forward degradation gap we observe is not a model deficiency -- it is a property of the domain. Re-evaluation on contemporary data should be part of the deployment lifecycle.
\end{enumerate}

\section{Limitations}

\paragraph{Evaluation scale.} Our model evaluation corpus of 300 documents, while stratified, is modest in size. Results on minority classes (e.g., 7 instances of ``closed'' outcomes in the 273-document validated subset) have wide confidence intervals. The public benchmark dataset (Section~\ref{sec:public_dataset}) partially addresses this limitation with 14,452 decisions, though the model evaluation results reported in this paper are based on the 273-document validated subset.

\paragraph{Class imbalance.} The case outcome label distribution reflects the natural distribution in EDRSR, where ``granted'' constitutes approximately 80\% of decisions. While this is realistic, it limits our ability to assess minority-class performance and inflates overall accuracy for models that default to the majority class.

\paragraph{API-only evaluation.} All models were evaluated via the AWS Bedrock API, which provides no visibility into tokenizer vocabulary, model weights, or inference configuration. Fertility measurements rely on the API's reported token counts, which may include special tokens or system prompt overhead. We mitigated this by using consistent prompts across all models, but minor systematic biases cannot be ruled out.

\paragraph{Single prompt template.} We used a single Ukrainian-language prompt template per task. Performance may vary with prompt engineering, chain-of-thought prompting, or English-language instructions, avenues we leave for future work.

\paragraph{Non-reasoning mode.} All evaluations were conducted in standard (non-reasoning) inference mode with temperature set to 0. Several models in our evaluation support extended reasoning or ``thinking'' modes, most notably Nemotron Super~3, whose reasoning mode is a key architectural feature, and Qwen~3, which supports a thinking/non-thinking toggle. Reasoning mode introduces an internal chain-of-thought before producing the final answer, which may substantially improve performance on tasks requiring multi-step legal reasoning, such as case outcome classification. Our results therefore represent a lower bound on the capabilities of reasoning-capable models. An ablation comparing standard vs.\ reasoning mode, particularly for Nemotron Super~3 on the case outcome task where it already leads at 96.0\%, is an important direction for future work.

\paragraph{Temporal specificity.} The model versions accessed via Bedrock in April--May 2026 may differ from those available at other times or through other providers. Our results reflect the specific model endpoints available during the experiment window.

\paragraph{No fine-tuning.} We evaluate only zero-shot and few-shot settings. Fine-tuned models would likely show different performance patterns, particularly for the norm extraction task where the structured output format is critical.

\paragraph{No Ukrainian-specific neural baselines.} While we establish a classical TF-IDF baseline (Section~\ref{sec:temporal}), we do not include Ukrainian-specific neural models (e.g., XLM-R fine-tuned on Ukrainian legal corpora, or domain-specific models trained on EDRSR data). Such baselines would contextualize whether the 86--96\% zero-shot accuracy achieved by general-purpose foundation models is competitive with, or still below, purpose-built alternatives.

\paragraph{Outcome label provenance.} Our outcome labels, while validated through a three-source majority vote, rely on rule-based extraction from the dispositive section. Documents with atypical structure (e.g., interlocutory orders, procedural rulings) were disproportionately excluded during validation, potentially biasing the remaining dataset toward decisions with clear-cut outcomes. Additionally, Nemotron Super~3 served as one of the three voters in the majority-vote tiebreaker, creating a potential circularity with its role as an evaluated model. Our tiebreaker bias analysis (Section~\ref{sec:tiebreaker_bias}) shows that Nemotron's lead holds on the 205-document easy subset where it had no tiebreaker role (98.0\%, 201/205, tied for first), but we acknowledge that a fully independent tiebreaker (e.g., GPT-4 or Gemini) would eliminate this concern entirely.

\section{Conclusion}

We have presented a systematic evaluation of seven foundation models on Ukrainian legal text, measuring both tokenizer efficiency and downstream task performance. Our key findings are:

\begin{enumerate}[leftmargin=*]
    \item \textbf{NVIDIA Nemotron Super~3 (120B) is the best single model for Ukrainian legal text}, achieving the highest composite score (83.1) across all three tasks, including 96.0\% on case outcome classification and 98.9\% on case type. It outperforms Mistral Large~3 (675B total, 41B active per token), a model with 5.6$\times$ more total parameters and 3.4$\times$ more active parameters, at one-third the API cost (\$3.61 vs.\ \$10.99). A routed multi-model ensemble (Maverick for classification, Nemotron for outcome, Llama~3.3 for extraction) achieves an even higher composite (85.1) at 37\% lower cost than Nemotron alone.

    \item \textbf{Tokenizer fertility varies by 1.6$\times$} across models on Ukrainian legal text, with Llama-family tokenizers (2.43--2.65 tokens/word) substantially more efficient than Qwen tokenizers (3.90 tokens/word). This directly affects API cost and effective context length: Qwen models consume 60\% more tokens per document than Llama models for identical input.

    \item \textbf{Few-shot prompting is counterproductive} for most models on Ukrainian legal classification tasks. A stratified few-shot ablation confirms that even distribution-matched examples degrade performance by up to 26 percentage points, ruling out example selection bias and implicating morphological interference intrinsic to Ukrainian-language demonstrations.

    \item \textbf{Temporal drift dominates model selection.} A cross-temporal generalization experiment on 15{,}000 documents across three epochs (pre-war, hybrid war, full-scale invasion) reveals that classifiers trained on pre-war data lose 27.9~pp when applied to full-scale era cases -- a degradation larger than the gap between the best and worst foundation models on any single task. The forward--backward asymmetry (backward transfer exceeds forward by 14.6~pp) suggests that legal language is additive: newer legal frameworks subsume older ones, but the reverse does not hold.

    \item \textbf{Systematic model selection via managed APIs is inexpensive.} The total cost of the core evaluation (7 models $\times$ 3 tasks $\times$ 2 modes $\times$ 273--300 documents) was \$31.41 (Table~\ref{tab:costs}). Including ablation studies and the temporal generalization experiment, the total cost was approximately \$70, demonstrating that comprehensive language-specific benchmarking is feasible even for resource-constrained teams.
\end{enumerate}

These findings underscore the importance of language-specific and \emph{temporally aware} evaluation before model deployment. English-language benchmarks, parameter counts, and static test sets are poor proxies for performance on morphologically rich, Cyrillic-script languages operating under evolving legal frameworks. For practitioners: Nemotron Super~3 offers the best accuracy--cost tradeoff for Ukrainian legal NLP; Llama~4 Maverick provides the cheapest inference at near-top accuracy; zero-shot prompting should be preferred over few-shot for Ukrainian; and temporal re-evaluation should be budgeted as part of the deployment lifecycle. We release our evaluation methodology, results, and the temporally annotated benchmark dataset to support practitioners building legal NLP systems for Ukrainian and related languages.

\paragraph{Data and code availability.} The evaluation code and aggregated results are available at \url{https://github.com/overthelex/rlhf-signals}. The public benchmark dataset (14,452 decisions, 2008--2026, seven outcome labels, three temporal epochs) is available at \url{https://huggingface.co/datasets/overthelex/ukrainian-court-decisions} (config: \texttt{case\_outcome\_temporal}). Individual court decisions are publicly available via the EDRSR API (\url{https://reyestr.court.gov.ua}).

\section*{Acknowledgments}

This work was conducted as part of the LEX AI platform development at legal.org.ua. LEX AI LLC is a member of the NVIDIA Inception program for AI startups. Compute costs for all experiments were covered by an AWS Activate grant (\$25,000 in AWS credits); no compute credits or other support was received from NVIDIA or any other model provider evaluated in this study. We thank the EDRSR for providing open access to court decisions, AWS for the Bedrock API infrastructure, and NVIDIA, Meta, Qwen, Mistral AI, and Amazon for making their foundation models accessible for independent evaluation.

\paragraph{Conflict of interest disclosure.} The author has no financial relationship with NVIDIA beyond membership in the NVIDIA Inception program, which provides business resources but did not fund or influence this research. All experiments were conducted on AWS infrastructure funded by an AWS grant. The evaluation methodology, model selection, and conclusions were determined independently. NVIDIA Nemotron Super~3's top ranking in our evaluation is an empirical finding, not a sponsored result.


\appendix

\section{Prompt Templates}
\label{app:prompts}

\subsection{Case Type Classification (Zero-Shot)}

\begin{quote}\small\ttfamily
\foreignlanguage{ukrainian}{Визнач тип судової справи з тексту рішення.
Відповідай ОДНИМ словом: цивільна, кримінальна,
господарська, або адміністративна.}\\[6pt]
\foreignlanguage{ukrainian}{Текст рішення:}\\
\{document\_text\}\\[6pt]
\foreignlanguage{ukrainian}{Тип справи:}
\end{quote}

\subsection{Case Outcome Classification (Zero-Shot)}

\begin{quote}\small\ttfamily
\foreignlanguage{ukrainian}{Визнач результат розгляду справи з тексту рішення.
Відповідай ОДНИМ з варіантів: задоволено, відмовлено,
залишено без розгляду, частково задоволено, закрито.}\\[6pt]
\foreignlanguage{ukrainian}{Текст рішення:}\\
\{document\_text\}\\[6pt]
\foreignlanguage{ukrainian}{Результат:}
\end{quote}

\subsection{Norm Extraction (Zero-Shot)}

\begin{quote}\small\ttfamily
\foreignlanguage{ukrainian}{Витягни всі правові норми (закон + стаття),
на які посилається суд у цьому рішенні.
Поверни відповідь у форматі JSON масиву:}\\
{[}\{"law": "\foreignlanguage{ukrainian}{назва}",\\
\hspace*{1em}"article": "\foreignlanguage{ukrainian}{номер}"\}{]}\\[6pt]
\foreignlanguage{ukrainian}{Текст рішення:}\\
\{document\_text\}\\[6pt]
\foreignlanguage{ukrainian}{Норми (JSON):}
\end{quote}

\section{Full Per-Model Results}
\label{app:full_results}

Table~\ref{tab:full_results} presents the complete results matrix for all model--task--mode combinations.

\begin{table}[h]
\centering
\caption{Complete results for all 42 model--task--mode combinations (7 models $\times$ 3 tasks $\times$ 2 modes). All metrics reported on the $n{=}273$ validated subset.}
\label{tab:full_results}
\small
\begin{tabular}{llR{1.5cm}R{1.5cm}R{1.5cm}}
\toprule
\textbf{Model} & \textbf{Mode} & \textbf{Case Type} & \textbf{Outcome} & \textbf{Norm F1} \\
\midrule
\multirow{2}{*}{Llama 4 Maverick} & Zero-shot & 98.9 & 91.2 & 0.487 \\
 & Few-shot & 92.7 & 91.9 & 0.488 \\
\midrule
\multirow{2}{*}{Llama 3.3 70B} & Zero-shot & 94.5 & 89.7 & 0.604 \\
 & Few-shot & 96.0 & 81.0 & 0.606 \\
\midrule
\multirow{2}{*}{Mistral Large 3} & Zero-shot & 95.6 & 91.6 & 0.561 \\
 & Few-shot & 94.9 & 85.3 & 0.575 \\
\midrule
\multirow{2}{*}{Nemotron Super 3} & Zero-shot & 98.9 & 96.0 & 0.543 \\
 & Few-shot & 94.5 & 83.2 & 0.564 \\
\midrule
\multirow{2}{*}{Nova Pro} & Zero-shot & 98.2 & 92.3 & 0.575 \\
 & Few-shot & 92.3 & 92.7 & 0.585 \\
\midrule
\multirow{2}{*}{Qwen3 235B} & Zero-shot & 97.4 & 93.8 & 0.463 \\
 & Few-shot & 98.5 & 67.8 & 0.476 \\
\midrule
\multirow{2}{*}{Qwen3 32B} & Zero-shot & 95.2 & 86.8 & 0.514 \\
 & Few-shot & 95.2 & 89.7 & 0.529 \\
\bottomrule
\end{tabular}
\end{table}

\section{Dataset Statistics}
\label{app:dataset}

\begin{table}[h]
\centering
\caption{Evaluation corpus statistics.}
\label{tab:dataset_stats}
\begin{tabular}{lr}
\toprule
\textbf{Statistic} & \textbf{Value} \\
\midrule
Total documents & 300 \\
Documents per case type & 75 \\
Outcome-validated subset & 273 \\
Excluded (label disagreement) & 27 \\
Case outcome: granted (\textit{zadovoleno}) & 230 / 273 (84.2\%) \\
Case outcome: denied (\textit{vidmovleno}) & 15 / 273 (5.5\%) \\
Case outcome: left without consideration & 21 / 273 (7.7\%) \\
Case outcome: closed (\textit{zakryto}) & 7 / 273 (2.6\%) \\
Case outcome: partially granted & 0 (originally 10, all disputed and excluded) \\
Source & EDRSR \\
Language & Ukrainian \\
Tokenizer fertility samples & 100 (6,000 chars each) \\
\bottomrule
\end{tabular}
\end{table}

\end{document}